# CASCADED FEATURE WARPING NETWORK FOR UNSUPERVISED MEDICAL IMAGE REGISTRATION

*Liutong Zhang, Lei Zhou, Ruiyang Li, Xianyu Wang, Boxuan Han, Hongen Liao*

Department of Biomedical Engineering, School of Medicine, Tsinghua University, Beijing, China

## ABSTRACT

Deformable image registration is widely utilized in medical image analysis, but most proposed methods fail in the situation of complex deformations. In this paper, we present a cascaded feature warping network to perform the coarse-to-fine registration. To achieve this, a shared-weights encoder network is adopted to generate the feature pyramids for the unaligned images. The feature warping registration module is then used to estimate the deformation field at each level. The coarse-to-fine manner is implemented by cascading the module from the bottom level to the top level. Furthermore, the multi-scale loss is also introduced to boost the registration performance. We employ two public benchmark datasets and conduct various experiments to evaluate our method. The results show that our method outperforms the state-of-the-art methods, which also demonstrates that the cascaded feature warping network can perform the coarse-to-fine registration effectively and efficiently.

## 1. INTRODUCTION

Deformable image registration plays an important role in many medical analysis tasks such as image fusion, organ atlas creation, and tumor growth monitoring [1]. The non-linear correspondence is established by predicting a dense deformation field. Traditional registration methods seek to estimate the deformation field by minimizing the cost function in associating with similarity metrics. However, these methods are usually computationally expensive and time-consuming.

Deep learning-based registration methods have been widely studied to achieve accurate and fast registration. These methods can be classified into supervised learning and unsupervised learning. The supervised learning-based methods are highly dependent on the ground-truth deformation field obtained by either traditional algorithms [2] or simulated deformations [3]. However, the registration performance is severely limited by the quality of the generated training data.

The unsupervised learning-based registration methods have drawn more and more attention because they don't need the ground truth during the training process. These methods only require a similarity measure function between the warped moving image and the fixed image, while the gradients can backpropagate through the differentiable warping operation [4]. Among them, Balakrishnan et al. [5] introduced a convolution neural network (CNN) named VoxelMorph (VM) to predict the deformation field. Kuang et al. [6] designed a new regularization function to improve the registration result. A probabilistic multilayer regularization network proposed by Liu et al. [7] adopted direct regularizations to the hidden layers. However, the above methods are not effective when dealing with complicated deformations especially with large displacements. To address this issue, the study [8] presented a coarse-to-fine framework by stacking multiple CNNs, but the networks were optimized separately. Zhao et al. [9] developed an end-to-end recursive cascaded network to enable all cascades to learn progressive registration cooperatively. However, the registration performance only increases significantly in the initial several cascades, which is not improved linearly with the number of cascades. Moreover, the recursive cascaded network introduces more parameters and consumes tremendous extra GPU memory, which is highly inefficient and limits its application to larger 3D images.

Inspired by recent approach on optical flow estimation [10], which handles the large displacements in the feature space, we propose a cascaded feature warping network to perform the coarse-to-fine registration. The proposed feature warping registration module can not only reduce the feature correspondence distance but also construct a matching cost in the feature space, and then these features are utilized to estimate the deformation field accurately. The multi-scale deformation field is generated by cascading the module from the bottom level to the top level. Furthermore, the multi-scale loss strategy is also applied to train the network. Experimental results on two benchmark datasets demonstrate that our method outperforms state-of-the-art methods by a large margin.

## 2. METHOD

The goal of medical image registration is to find the optimal deformation filed $\varphi$, which can accurately warp the moving image $I_m$ to align with the fixed image $I_f$. Here we propose a novel cascaded feature warping network to achieve the coarse-to-fine registration. As illustrated in Fig.1(a), the unaligned images are first imported to a shared-weights encoder network to obtain the feature pyramids. Then the cascaded feature warping registration is applied to get the multi-scale deformation field. Besides, the feature warping registration module reduces the correspondence distance by warping the moving feature and constructs a matching cost through the correlation layer as in Fig.1(b). Moreover, the network is optimized by the multi-scale loss.

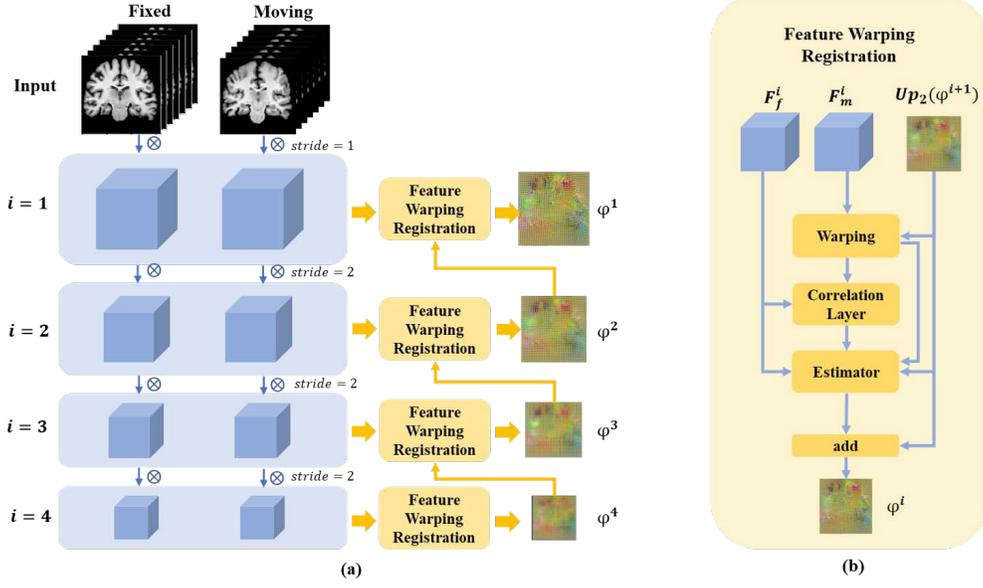

**Fig.1.** Illustrations of (a) cascaded feature warping network (b) feature warping registration module

**2.1. Feature Pyramid**

Given two input images $I_m$ and $I_f$, the shared-weights encoder network is used to generate the corresponding feature pyramids $F_m^i$ and $F_f^i$ ($i = 1,2 \cdots N$) as in Fig.1(a). Each level of the encoder contains two convolution blocks, and the convolution block is a $3^3$ 3D convolution layer followed by a leaky rectified linear unit (LeakyReLU) activation with a negative slope of 0.1. The stride of the first convolution block is set to 2 except for the top pyramid level. Consequently, the features at the $i$th level are $1/2^{i-1}$ of the input image size.

**2.2. Feature Warping Registration**

*2.2.1. Moving Feature Warping*

Given the moving feature $F_m^i$ and fixed feature $F_f^i$ at the $i$th level, it's challenging to predict the accurate deformation field without other strategies. The reason is that the convolutional layer can only capture the correspondence in the local receptive field. To reduce the spatial correspondence distance of these two features, we use the deformation field predicted in the previous level to warp the moving feature.

Specifically, the deformation field $\varphi^{i+1}$ is firstly upsampled with a factor of 2 and then adopted to warp the moving feature $F_m^i$. The differentiable warping operation is implemented by bilinear interpolation according to [4]. The whole process can be written as:

$$F_w^i = F_m^i \circ up_2(\varphi^{i+1}) \quad (1)$$

where $F_w^i$ is the warped feature at the $i$th level.

*2.2.2. Correlation Layer*

To measure the matching cost of associating a voxel with its corresponding voxels at another image, we use the features to construct a cost volume. The matching cost is defined as the correlation [10] between the fixed feature and the warped feature:

$$F_{corr}^i(x_1) = \frac{1}{C}(F_f^i(x_1))^T F_w^i(x_2), |x_1 - x_2|_\infty \leq d \quad (2)$$

where $T$ represents transpose operator, $C$ is the channel number of the feature $F_f^i$ and $d$ is the search range. The dimension of the 4D matching cost $F_{corr}^i$ is $(2d + 1)^3 \times D^i \times H^i \times W^i$, where $D^i$, $H^i$ and $W^i$ denote the depth, height, and width of the $i$th pyramid level. Considering that one-voxel movement at the bottom level corresponds to $2^{N-1}$ voxels at the full resolution image, the $d$ is set to a low value.

*2.2.3. Deformation Field Estimator*

The deformation field estimator consists of multiple convolution blocks. The inputs of the estimator include the matching cost $F_{corr}^i$, the fixed feature $F_f^i$, the warped feature $F_w^i$ and the upsampled deformation field $up_2(\varphi^{i+1})$. Considering that the moving feature has been warped, our estimator is adopted to predict the residual deformation field. The final deformation field is the estimator result plus the upsampled deformation field. The whole process can be described as:

$$\varphi^i = Conv\left(F_{corr}^i, F_f^i, F_w^i, up_2(\varphi^{i+1})\right) + up_2(\varphi^{i+1}) \quad (3)$$

where $Conv$ denotes the convolution blocks of the estimator.

Table 1. Dice score of different methods on Mindboggle101|LPBA40. The last column is the network parameters size.

| Method | Frontal | Parietal | Occipital | Temporal | Cingulate | Putamen | Hippo | Params(K) |
|---|---|---|---|---|---|---|---|---|
| SyN [13] | 0.520\|0.703 | 0.452\|0.656 | 0.402\|0.653 | 0.544\|0.694 | 0.630\|0.724 | −\|0.762 | −\|0.736 | − |
| PMRN [7] | 0.579\|0.711 | 0.559\|0.661 | 0.430\|0.660 | 0.544\|0.679 | 0.546\|0.691 | −\|0.711 | −\|0.701 | − |
| VM [5] | 0.585\|0.687 | 0.525\|0.612 | 0.443\|0.582 | 0.587\|0.649 | 0.672\|0.679 | −\|0.736 | −\|0.709 | 394.7 |
| RCN [9] | 0.624\|0.711 | 0.564\|0.644 | 0.489\|0.629 | 0.622\|0.702 | 0.689\|0.729 | −\|0.776 | −\|0.736 | 1189.4 |
| VM×2 | 0.613\|0.697 | 0.553\|0.622 | 0.473\|0.592 | 0.615\|0.663 | 0.696\|0.705 | −\|0.758 | −\|0.729 | 1576.8 |
| DB-VM | 0.604\|0.692 | 0.544\|0.619 | 0.462\|0.587 | 0.605\|0.660 | 0.689\|0.697 | −\|0.754 | −\|0.721 | 638.7 |
| 2-Cas-VM | 0.629\|0.702 | 0.570\|0.630 | 0.493\|0.608 | 0.626\|0.680 | 0.704\|0.715 | −\|0.776 | −\|0.731 | 792.9 |
| Baseline-1 | 0.594\|0.697 | 0.536\|0.622 | 0.453\|0.594 | 0.595\|0.660 | 0.682\|0.691 | −\|0.744 | −\|0.710 | 629.4 |
| Baseline-2 | 0.643\|0.717 | 0.586\|0.653 | 0.528\|0.643 | 0.636\|0.707 | 0.710\|0.729 | −\|0.776 | −\|0.747 | 629.4 |
| Ours | **0.652\|0.720** | **0.601\|0.669** | **0.538\|0.666** | **0.645\|0.717** | **0.715\|0.737** | **−\|0.784** | **−\|0.757** | 667.4 |

### 2.3. Multi-scale loss

We also use the multi-scale similarity loss to enhance the coarse-to-fine registration. To ensure the smoothness of the deformation field, an additional regularization term is also adopted at each level. We formulate the similarity loss by using negative local cross correlation (NLCC) and encourage a smooth deformation field using a diffusion regularization on its spatial gradients. The complete loss is therefore characterized as:

$$L = \sum_{i=1}^{N} \frac{1}{2^{i-1}} \left( NLCC(I_m^i \circ \varphi^i, I_f^i) + \lambda \|\nabla \varphi^i\|_2^2 \right) \quad (4)$$

where $I_m^i, I_f^i$, and $\varphi^i$ denote the downsampled moving image, the downsampled fixed image, and the estimated deformation field at the $i$th level, and $\lambda$ is the regularization parameter. Besides, a higher level is assigned to lower loss weight and the window size of NLCC also linearly decreases from the top level to the bottom level.

## 3. EXPERIMENTS AND RESULTS

We evaluate the proposed method on 3D brain MRI registration with two public datasets, Mindboggle101 [11] and LPBA40 [12]. Mindboggle101 contains 101 T1-weighted MR images, which are annotated with 25 cortical regions. LPBA40 consists of 40 T1-weighted MR images and each volume has a segmentation mask with 56 anatomical labels. For a fair comparison, experiments are conducted following the recent work [7]. 42 images with 1722 pairs are used for training and 20 images with 380 pairs are used for testing on Mindboggle101. The first 30 images with 870 pairs are adopted as the training set and the remaining 10 images with 90 pairs are adopted as the testing set on LPBA40. Besides, standard preprocessing steps, including spacing re-sampling (1mm), spatial normalization, and rescaling to [0,1], are performed, and the resulting images are center cropped to $160 \times 192 \times 160$. To evaluate the registration performance, we measure the overlap between the warped moving segmentation mask and the fixed image segmentation mask using Dice metric.

The network is implemented with PyTorch and trained by the Adam optimizer with a learning rate of 0.0001 for 15000 steps. The regularization parameter $\lambda$ is set to 1 according to [5]. Batch size and the search range d is also set to 1 due to the limitation of GPU memory.

### 3.1. Multi-scale Registration Visualization

We first visualize the multi-scale deformation field and registration results in Fig.2. We can find that the network gradually achieves more accurate registration with the enrichment of detailed information from the bottom level to the top level. It also demonstrates that our cascaded feature warping network implements the coarse-to-fine registration effectively.

### 3.2. Comparison with Other Methods

We further compare our method with several unsupervised registration approaches, including SyN [13], VM [5], RCN [9], and PMRN [7]. We follow [7] and report the average Dice value on five or seven large regions which are grouped from the initial small regions. Table 1 shows the results on the two benchmark datasets. The results imply that our method outperforms the others by a large margin. Fig.3 also presents the registration results generated by different methods. These all indicate that our network can conduct the registration accurately.

In consideration of the fact that the parameters of our network increase compared with VM, we also compare our method with three VM variants VM×2, DB-VM, and 2-Cas-VM. VM×2 doubles the feature channels of every convolutional layer and DB-VM doubles the number of convolutional layers at each level. 2-Cas-VM follows [9] to cascade two subnetworks. The results on the two datasets are shown in Table 1. VM×2 and DB-VM perform better than the original VM but worse than 2-Cas-VM. However, our method can obtain the best performance with a relatively small amount of parameters. This experiment implies that our improvements are essentially based on the proposed cascaded feature warping registration framework rather than simply introducing more parameters.

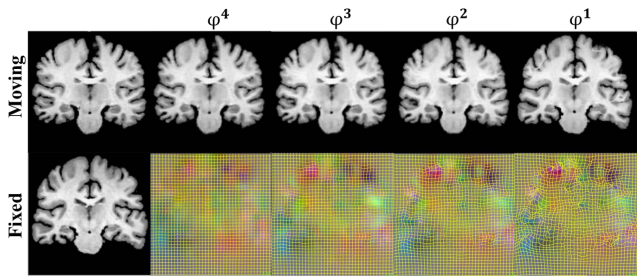

**Fig.2.** Multi-scale deformation field and registration results

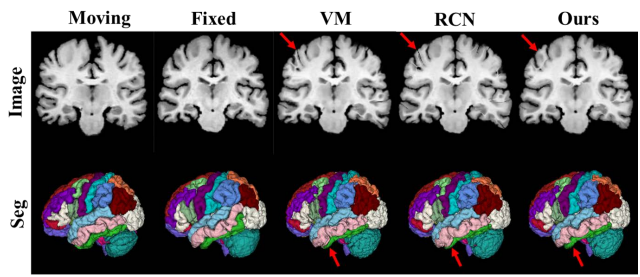

**Fig.3.** Registration results generated by different methods

### 3.3. Ablation Study

We also conduct an ablation study to verify the effectiveness of the main modules in our network. We construct two baselines. The first one (Baseline-1) is to remove the moving feature warping and the correlation layer based on our network. The second one (Baseline-2) is to exclude the correlation layer. The comparison results are also shown in Table 1. We can observe that the moving feature warping greatly improves the registration performance and the correlation layer leads to further improvement.

### 4. CONCLUSION

This paper presents a cascaded feature warping network for unsupervised medical image registration. We design a feature warping registration module to estimate the deformation field accurately at each pyramid level. The coarse-to-fine registration is performed by cascading the module from the bottom level to the top level. The multi-scale loss is also introduced to train the network. Experimental results on two benchmark datasets show that our method achieves better performance and outperforms other approaches.

### 5. COMPLIANCE WITH ETHICAL STANDARDS

This research study was conducted using human subject data made available in open access. Ethical approval was not required as confirmed by the license attached with the open access data.

### 6. ACKNOWLEDGMENTS

The authors acknowledge supports from National Natural Science Foundation of China (82027807, 81771940), and Beijing National Science Foundation (7212202).

### 7. REFERENCES

[1] G. Haskins, U. Kruger, and P. Yan, "Deep learning in medical image registration: a survey," Machine Vision and Applications, vol. 31, no. 1, p. 8, 2020.
[2] X. Cao, J. Yang, J. Zhang, D. Nie, M. Kim, Q. Wang, and D. Shen, "Deformable image registration based on similarity-steered CNN regression," in MICCAI, Springer, pp. 300-308, 2017.
[3] J. Krebs et al, "Robust non-rigid registration through agent-based action learning," in MICCAI, Springer, pp. 344-352, 2017.
[4] M. Jaderberg, K. Simonyan, A. Zisserman, and K. Kavukcuoglu, "Spatial transformer networks," in NIPS, pp. 2017-2025, 2015.
[5] G. Balakrishnan, A. Zhao, M. R. Sabuncu, J. Guttag, and A. V. Dalca, "An unsupervised learning model for deformable medical image registration," in CVPR, pp. 9252-9260, 2018.
[6] D. Kuang, and T. Schmah. "Faim–a convnet method for unsupervised 3d medical image registration," in MLMI, Springer, pp. 646-654, 2019.
[7] L. Liu, X. Hu, L. Zhu, and PA. Heng, "Probabilistic multilayer regularization network for unsupervised 3D brain image registration," in MICCAI, Springer, pp. 346-354, 2019.
[8] B. D. de Vos, F. F. Berendsen, M. A. Viergever, H. Sokooti, M. Staring, and I. Išgum, "A deep learning framework for unsupervised affine and deformable image registration," Medical Image Analysis, vol. 52, pp. 128–143, 2019.
[9] S. Zhao, Y. Dong, EI. Chang, and Y. Xu, "Recursive cascaded networks for unsupervised medical image registration." in ICCV, pp. 10600-10610, 2019.
[10] A. Dosovitskiy, P. Fischer, E. Ilg, P. Hausser, C. Hazirbas, and V. Golkov, "FlowNet: Learning optical flow with convolutional networks," in ICCV, pp. 2758–2766, 2015.
[11] A. Klein and J. Tourville, "101 Labeled Brain Images and a Consistent Human Cortical Labeling Protocol," Frontiers in neuroscience, vol. 6, 2012.
[12] D. W. Shattuck et al, "Construction of a 3D probabilistic atlas of human cortical structures," NeuroImage, vol. 39, no. 3, pp. 1064–1080, 2008.
[13] B. B. Avants, C. L. Epstein, M. Grossman, and J. C. Gee, "Symmetric diffeomorphic image registration with cross-correlation: Evaluating automated labeling of elderly and neurodegenerative brain," Medical Image Analysis, vol. 12, no. 1, pp. 26–41, 2008.